\documentclass[11pt,a4paper]{article}
\usepackage[hyperref]{eacl2021}
\usepackage{times}
\usepackage{latexsym}

\usepackage{microtype}
\usepackage{caption,subcaption}
\newcommand{\STAB}[1]{\begin{tabular}{@{}c@{}}#1\end{tabular}}

\usepackage{amsmath,amssymb,dsfont}
\usepackage{enumitem}
\usepackage{graphicx}
\usepackage[normalem]{ulem}
\usepackage{url}

\usepackage[linesnumbered,ruled,vlined,algo2e]{algorithm2e}

\usepackage{microtype}

\usepackage{multicol}
\usepackage{multirow}
\usepackage{array}
\usepackage{tabu,colortbl}
\usepackage{hhline}
\usepackage{subcaption}

\newcommand{\comment}[1]{}

\usepackage{xcolor}
\usepackage{transparent}

\usepackage{fancybox} 
 \usepackage{microtype}
\usepackage{amsmath}
\usepackage{amssymb}
\usepackage{bbm}
\usepackage{graphicx}
\usepackage{multicol}
\usepackage{tabularx}
\usepackage{booktabs}
\usepackage{algorithmic}
\usepackage{algorithm}
\usepackage{cleveref}
\usepackage{multirow}
\usepackage{tabularx}

\usepackage{xspace} 

\usepackage{bm}

\definecolor{our_blue}{rgb}{0.21747533000128158, 0.5305292836088684, 0.7548225041650647}
\definecolor{our_red}{rgb}{0.7364705882352941, 0.08, 0.10117647058823528}

\crefformat{section}{\S#2#1#3}
\crefformat{subsection}{\S#2#1#3}
\crefformat{subsubsection}{\S#2#1#3}

\aclfinalcopy 
\def\aclpaperid{316} 

\title{Does the Order of Training Samples Matter? Improving Neural Data-to-Text Generation with Curriculum Learning}

\author{ 
  Ernie Chang, Hui-Syuan Yeh, Vera Demberg \\
  Dept. of Language Science and Technology, Saarland University \\
    {\tt \{cychang,yehhui,vera\}@coli.uni-saarland.de}
  \\ 
}

\date{}

\begin{document}
\maketitle
\begin{abstract}

Recent advancements in data-to-text generation largely take on the form of neural end-to-end systems. 
Efforts have been dedicated to improving text generation systems by changing the order of training samples in a process known as \emph{curriculum learning}.
Past research
on sequence-to-sequence learning showed that \emph{curriculum learning} helps to improve both the performance and convergence speed. 
In this work, we delve into the same idea surrounding the training samples consisting of structured data and text pairs, where at each update, the curriculum framework selects training samples based on the model's competence.
Specifically, we experiment with various difficulty metrics and put forward a \emph{soft edit distance metric}
for ranking training samples.
Our benchmarks show faster convergence speed where training time is reduced by $38.7$\% and performance is boosted by $4.84$ BLEU.

\end{abstract}

\section{Introduction}

Neural data-to-text generation has been the subject of much recent research. 
The task aims at transforming source-side structured data into target-side natural language descriptions~\cite{reiter2000building,barzilay-lapata-2005-modeling}. 
The process typically involves mini-matches which are randomly sampled with a fixed size from the training set to feed into the model at each training step.
In this paper, we apply curriculum learning to this process, which was explored in neural machine translation~\cite{Platanios2019,zhou2020uncertainty}, and show how it can help in neural data-to-text generation.

The main idea in curriculum learning is to present the training data in a specific order, starting from {\em easy} examples and moving on to more {\em difficult} ones, as the learner becomes more {\em competent}. 
When starting out with easier instances, the risk of getting stuck in local optima early on in training is reduced, since the loss functions in neural models are typically highly non-convex~\cite{bengio2009curriculum}.
This learning paradigm enables flexible batch configurations by considering the material properties as well as the state of the learner. 
The idea brings in two potential benefits:
(1) It speeds up the convergence and reduces the computational cost. 
(2) It boosts the model performance, without having to change the model or add data. 

With the release of large data-to-text datasets (e.g.~Wikibio~\cite{lebret2016neural}, Totto~\cite{parikh2020totto}, E2E~\cite{novikova2017e2e}), neural data-to-text generation is now at a point where training speed and the order of samples may begin to make a real difference. 
We here show the efficacy of curriculum learning with a general LSTM-based sequence-to-sequence model and define \emph{difficulty metrics} that can assess the training instances, using a sucessful \emph{competence function} which estimates the model capability during training. Such metrics have not yet been explored in neural data-to-text generation.

In this paper, we explore the effectiveness of various difficulty metrics and propose a soft edit distance metric, which leads to substantial improvements over other metrics. 
Crucially, we observe that {difficulty metrics} that consider data-text samples jointly lead to stronger improvements than metrics that consider text or data samples alone.
In summary, this work makes the following contributions towards neural data-to-text generation:
\begin{enumerate}
    \item We show that by simply changing the order of samples during training, neural models can be improved via the use of curriculum learning. 
    \item We explore various \emph{difficulty metrics} at the level of the data, text, and data-text pairs, and propose an effective novel metric.
\end{enumerate}

\section{Related work}
\label{sec:related-work}

The idea of teaching algorithms in a similar manner as humans, incrementally from easy concepts to more difficult ones dates back to \emph{incremental learning}, which was discussed in light of theories of cognitive development relating to the processes of acquisition in young children~\cite{elman1993learning, krueger2009flexible,plunkett1993rote}.
\citet{bengio2009curriculum} first demonstrated empirically that curriculum learning approaches can decrease training times and improve generalization; later approach address these issues by changing the mini-batch sampling strategy to also include model competence~\cite{kocmi2017curriculum,zhou2020uncertainty,Platanios2019,liu2020norm,zhang2018empirical,zhang2019curriculum}.
While sample difficulty can be assessed for text samples and data samples or jointly, various measures have been proposed for text samples including n-gram frequency \citet{haffari2009machine,Platanios2019}, token rarity, and  sentence length~\cite{liu2020norm,Platanios2019}.
Our approach considers data and text jointly, similar to edit distance metric -- Levenshtein~\cite{levenshtein1966binary} and Damerau-Levenshtein Distance~\cite{damerau1964technique,brill2000improved}, which was used as a content ordering metric in ~\citet{wiseman2017challenges} to measure the extent of alignment between data slots and text tokens. 






\section{Preliminaries of Curriculum Learning}
We base our curriculum learning framework on the two standard components: 
(1) model competence (how capable the current model is at time $t$), and
(2) sample difficulty, which makes independent judgement on each sample's difficulty. 
Specifically, we adopt the competence function $c(t)$ for a model at time $t$ as in~\citet{Platanios2019,liu2020norm}:
\begin{equation}
	c_{\textrm{sqrt}}(t) \in (0,1] = \min\left( 1, \sqrt{t \frac{1 - c_0^2}{\lambda_t} + c_0^2} \right).
	\label{eq:sqrt_competence}
\end{equation}
Where $\lambda_t$ is a hyperparameter defining the length of the curriculum and is set to $2.5$ as in~\citet{liu2020norm}. $c_0 = 0.1$ as~\citet{Platanios2019}. 
Following this formulation, the number of new training examples per unit time is reduced as training progresses to give the learner sufficient time to obtain new knowledge. 
The sequence-to-sequence model learns using the curriculum as outlined in Algorithm~\ref{alg:ccl} by primarily making batch-wise decisions about which samples to add to each batch. 
This decision is determined by comparing the \emph{competence} score with the \emph{difficulty} score as shown in Algorithm~\ref{alg:ccl}. 



\begin{algorithm2e}[t]
\caption{Curriculum Learning Algorithm}
\label{alg:ccl}
\small
\KwIn{Training set, $\mathcal{D}=\{s_d, s_t\}_{i=1}^M$, consisting of $M$ samples, model ($\mathcal{T}$), difficulty metric ($d$), and competence function ($c$).}
Compute the difficulty, $d(s_i)$, for each data-text pair $\in \mathcal{D}$  (Section~\ref{sec:diff}). \\
Compute the CDF score $\bar{d}(s_i)$ of $d(s_i)$, where $\bar{d}(s_i) \in [0,1]$ (See Figure~\ref{fig:test}). \\
    \For{training step $t=1,\hdots$}{
    	Compute the model competence $c(t)$ with $\mathcal{T}$. \\
    	Train $\mathcal{T}$ on sampled data batch, $B_t$, drawn uniformly from all $s_i \in \mathcal{D}$, such that $\bar{d}(s_i) \leq c(t)$. \\
    	\If{$c(t)=1$}{
    	break. \\}
    }
\KwOut{Trained model.}
\end{algorithm2e}

\section{Difficulty Metrics}
\label{sec:diff}
For ease of discussion, we denote sequence to be $s$, which can be either data or text, or their concatenation. 
For comparison, the difficulty metrics use the unit tokens as tokenized by
SpaCy\footnote{ \url{https://spacy.io/api/tokenizer} }.
We begin with discussion on \emph{length} and \emph{word rarity}, which were previously applied by~\citet{kocmi2017curriculum,Platanios2019} on text sentences.

\begin{figure*}[t]
  \centering
\includegraphics[width=0.95\textwidth]{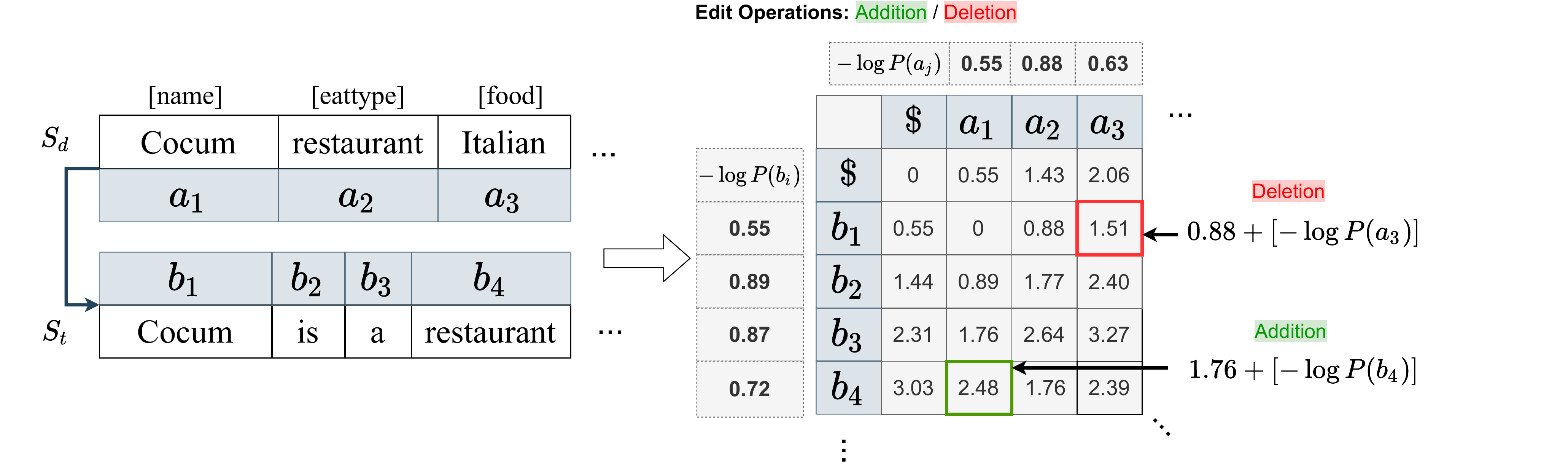}
\caption{\small Depiction of the process of \emph{soft edit distance metric} with the \emph{Wagner-Fischer table}. Each cell in the table represents the edit distance to convert data substring's into the text substring.
}
\label{fig: example}
\end{figure*}

\paragraph{Length.}
Length-based difficulty is based on the intuition that longer sequences are harder to encode, and that early errors may propagate during the decoding process, making longer sentences also harder to generate.
It is defined as:
\begin{align}
	d_{\textrm{length}}(s) &= N.
	\label{eq:sentence_length}
\end{align}

\paragraph{Rarity.}
 \emph{Word rarity} of a sentence is defined as the product of the unigram probabilities~\cite{Platanios2019}. 
 This metric implicitly incorporates information about sentence length since longer sentence scores are sum of more terms and are thus likely to be larger.
The difficulty metric for word rarity of a sequence $s$ is defined as:
\begin{align}
	d_{\textrm{rarity}}(s) &= -\sum_{k=1}^{N} \log p(w_k).
	\label{eq:sentence_length}
\end{align}

\begin{figure*}
\centering
\begin{minipage}{.45\textwidth}
  \begin{subfigure}{\linewidth}
    \centering
    \includegraphics[width=1.0\linewidth]{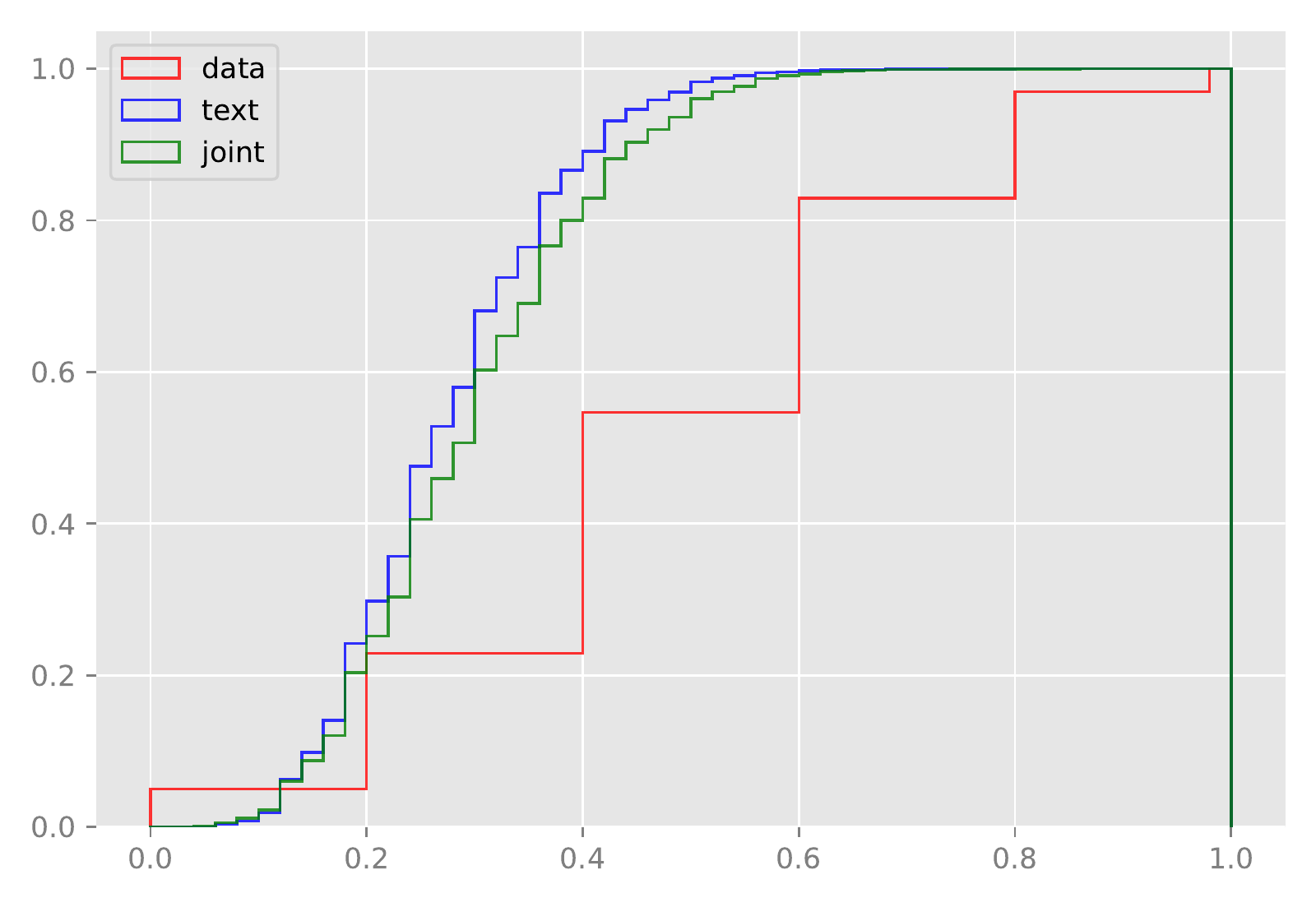}
    \caption{\emph{length}}
    \label{fig:sub1}
  \end{subfigure}\\[1ex]
  \begin{subfigure}{\linewidth}
    \centering
    \includegraphics[width=1.0\linewidth]{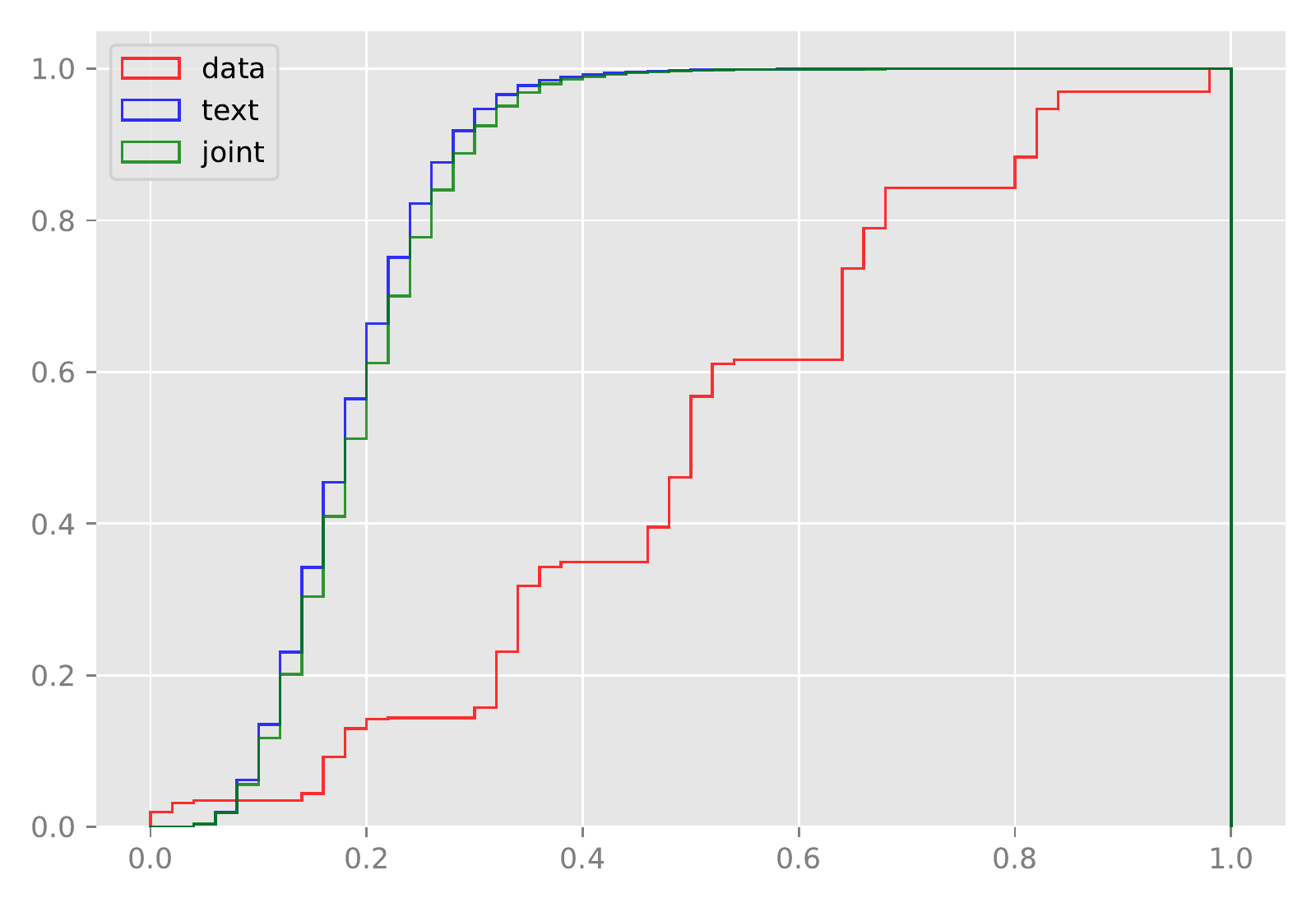}
    \caption{\emph{rarity}}
    \label{fig:sub2}
  \end{subfigure}
\end{minipage}%
\begin{minipage}{.45\textwidth}
  \begin{subfigure}{\linewidth}
    \centering
    \includegraphics[width=1.0\linewidth]{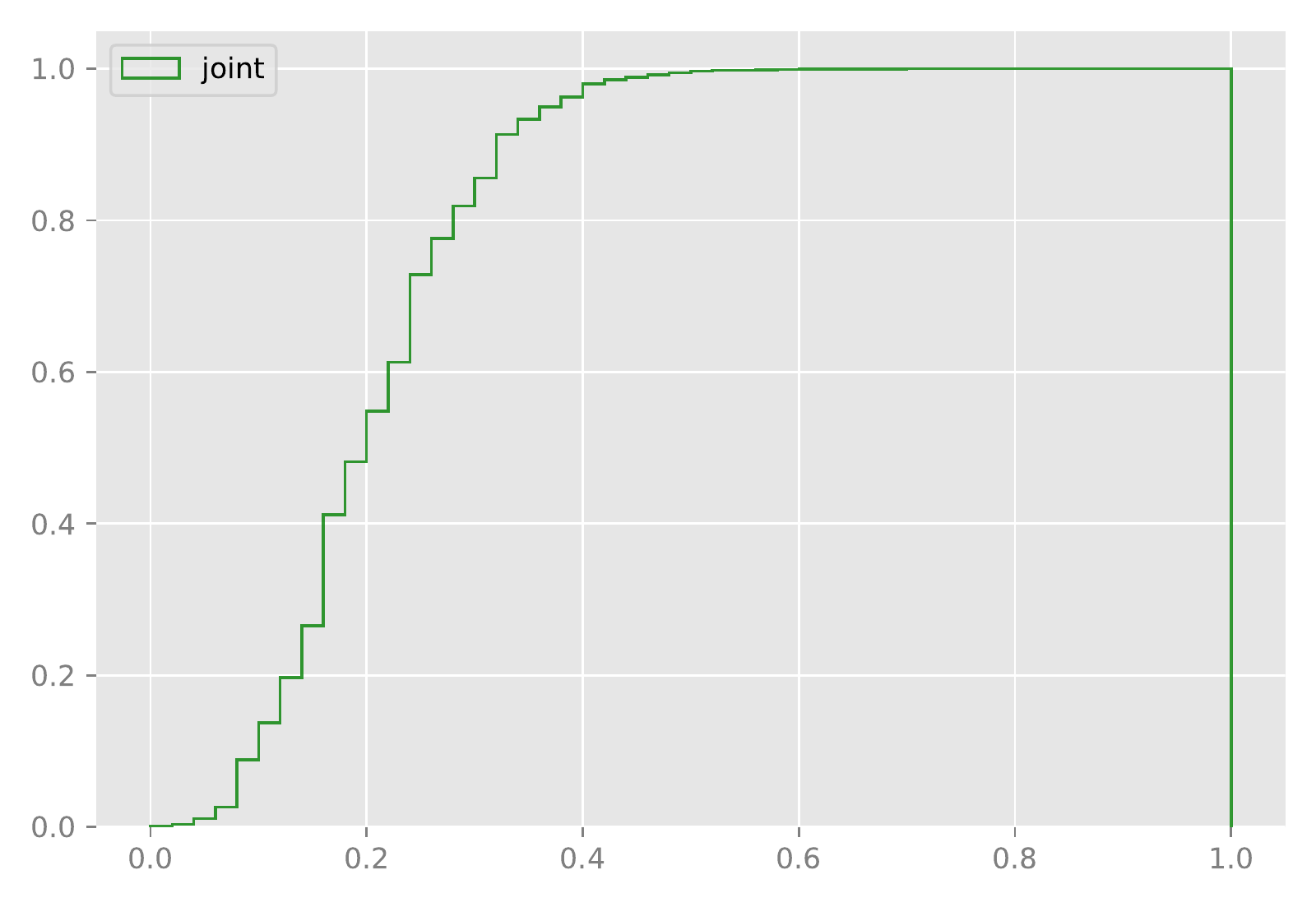}
    \caption{\emph{Damerau-Levenshtein distance}}
    \label{fig:sub4}
  \end{subfigure}\\[1ex]
  \begin{subfigure}{\linewidth}
    \centering
    \includegraphics[width=1.0\linewidth]{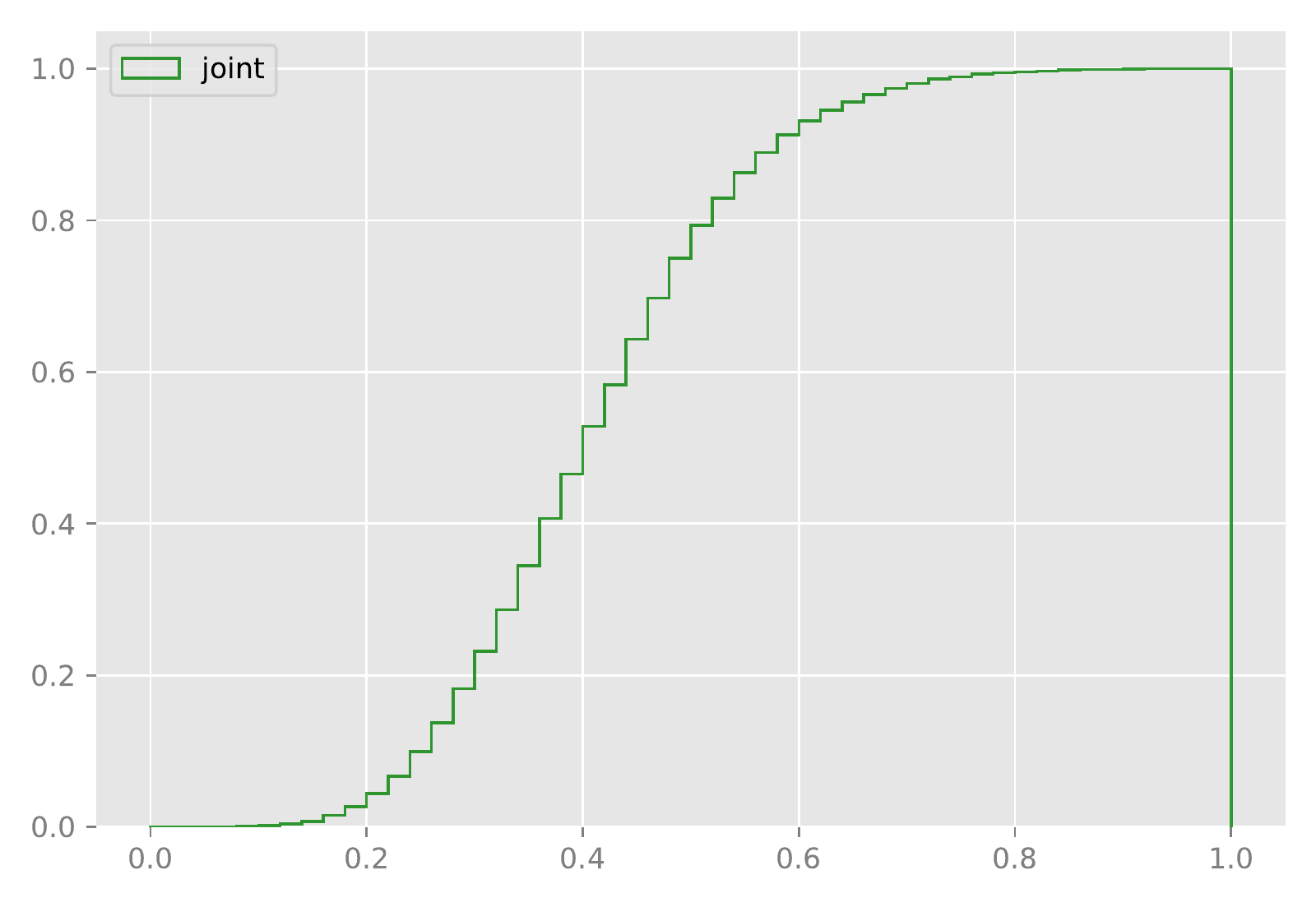}
    \caption{\emph{soft edit distance}}
    \label{fig:sub5}
  \end{subfigure}
\end{minipage}
\caption{The histogram of the cumulative density function for difficulty metrics.}
\label{fig:test}
\end{figure*}

\paragraph{Damerau-Levenshtein Distance.}

To consider data and text jointly, we measure the alignment between data slots and text using \emph{the Damerau-Levenshtein Distance} ($d_{\textrm{dld}}$)~\cite{brill2000improved}. 

We calculate the minimum number of edit operations needed to transform data ($s_{d}$) into text
($s_{t}$)\footnote{Previous work applies \emph{the Damerau-Levenshtein Distance} to slots in data (e.g.~``[name]'' in Figure~\ref{fig: example})  and extracts slots from text.}, and relies only four operations: (a) substitute a word in $s_{d}$ to a different word, (b) insert a word into $s_{d}$, (c) delete a word of $s_{d}$, and (d) transpose two adjacent words of $s_{d}$.
The process involves recursive calls that compute distance between substrings $s^{i}_{d} \in s_{d}$ and $s^{i}_{t} \in s_{t}$ at $i^{th}$ comparison.

\paragraph{Soft Data-to-Text Edit Distance.}
We here present the proposed  \emph{soft edit distance (SED)}: 
(1) We include the basic \emph{add} and \emph{delete} edit operations as in the Levenshtein Distance~\cite{levenshtein1966binary}, which was used in Levenshtein Transformer~\cite{gu2019levenshtein} as the only two necessary operations for decoding sequences since it correlates well with human text writing where humans ``\textit{can revise, replace, revoke or delete any part of their generated text}''.
We call this variant the plain edit distance \emph{(PED)}.
(2) Next, we weight the indicator function $\mathbbm{1}(s^{i}_{d},s^{i}_{t})$ for each edit operation with the negative logarithmic unigram probability $-\log p(w)$ for each token $w \in s^{i}_{ \{ t | d \} }  $, in order to incorporate the idea of \emph{word rarity} into the edit distance metric.
For the \emph{delete} operation, we use the $w \in s^{i}_{  d  }$ and for \emph{add} operations, we use $w \in s^{i}_{  t  }$.
This is unlike the previous proposal by \citet{brill-moore-2000-improved}, in which edits are weighted by the token transition probabilities -- this is not suitable for our scenario because there is no natural order of the slot sequence in data samples.

The soft distance metric $d_{\textrm{sed}}$ is in principle similar to calculating the logarithmic sum as defined in the \emph{rarity} function, but instead incrementally compares all substrings and calculates their edit distances. 
This way, $d_{\textrm{sed}}$ includes the information on \emph{length}, \emph{rarity} but also combining the edit operations. 
We show this process in Figure~\ref{fig: example}.


Note that we can compute \emph{length} and \emph{rarity} on the concatenation of input data and text sequence, or as individual sequences; whereas \emph{Damerau-Levenshtein distance} and \emph{soft edit distance} are computed jointly on data and text.

\begin{figure*}[t]
  \centering
\includegraphics[width=\textwidth]{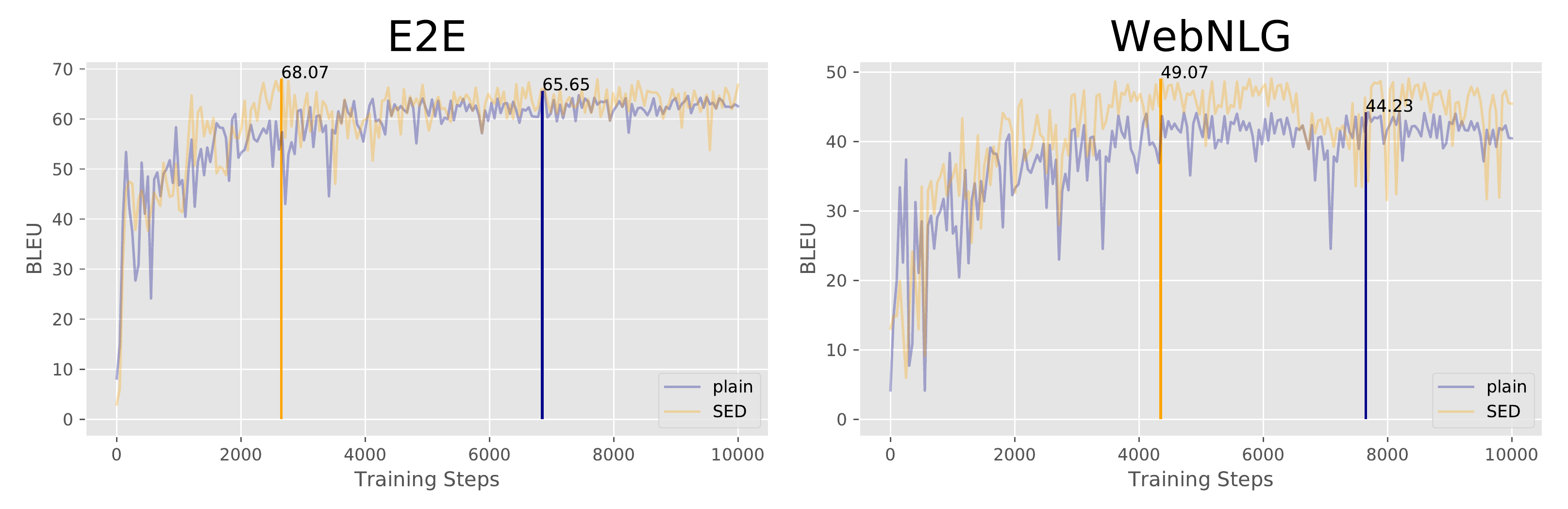}
\caption{\small A plot of performance (BLEU) versus the number of steps for E2E and WebNLG datasets. \textbf{Vertical bars} indicate where the maximum BLEU scores are attained for \textbf{plain} and \textbf{SED}.  
}
\label{fig:step}
\end{figure*}

\section{Experiment Setting}
\label{sec:exp}
\paragraph{Data.}
We conduct experiments on the E2E~\cite{novikova2017e2e} and WebNLG~\cite{colin-etal-2016-webnlg} datasets. 
E2E is a crowd-sourced dataset containing 50k instances in the restaurant domain. 
The inputs are dialogue acts consisting of three to $8$ slot-value pairs. 
WebNLG contains 25k instances describing entities belonging to $15$ distinct DBpedia categories, where the data contain are up to $7$ RDF triples of the form \emph{(subject, relation, object)}. 

\paragraph{Configurations.}
The LSTM-based model is implemented based on PyTorch~\cite{pytorch}.
We use $200$-dimensional token embeddings and the Adam optimizer with an initial learning rate at $0.0001$. 
Batch size is kept at $28$, and we decode with beam search with size $5$. 
The performance scores are averaged over $5$ random initialization runs. 

\paragraph{Settings.}

We first perform ablation studies (Table~\ref{tb:lmaug}) on the impact of difficulty metrics on data, text or both (joint).
We also analyse the average bin size for each metric -- a metric that gives the same score to many instances creates large bins. This means that the order of samples within the bin will still be random. On the other hand, a metric that assigns a lot of different difficulty scores to the instances can yield a more complete ordering (and a smaller step size in moving from one level of difficulty to the next). 
We present the change in performance (BLEU) as the training progresses in order to compare the various difficulty metrics on both datasets (See Figure~\ref{fig:step}). 

\section{Results \& Analysis}

\begin{table}[t!]
\centering
\resizebox{\columnwidth}{!}{
\begin{tabular}{cl|ccccccc}

 & \small\textbf{source}     & \textbf{\small plain}             & \textbf{\small L}  & \textbf{\small R}   &  \textbf{\small DLD } &  \textbf{\small PED} &  \textbf{\small SED}  \\ \hline
\multirow{3}{*}{\STAB{\rotatebox[origin=c]{90}{ \textbf{ \small BLEU} }}} & data   & \textbar  & 65.34   &  66.41  & -  &  -    \\ 
&text  & 65.65     & 65.77   &  67.01  & -  &  - &  -     \\
&joint & \textbar  & 66.21   &  67.29 & 66.58 &  66.26  &  \textbf{68.07}   \\ \hline
\multirow{3}{*}{\STAB{\rotatebox[origin=c]{90}{ \textbf{ \small Bin Size} }}} & data   & -   & 7010.17   &  385.88 & - & - &  -   \\ 
& text       & -  & 737.91   &  1.04 & - &  -  & -  \\
& joint      & -  & 25.0   &  1.04 & 32.26 & 21.74 &   \textbf{1.02}     \\ \hline
\multirow{3}{*}{\STAB{\rotatebox[origin=c]{90}{ \textbf{ \small Human} }}}
& fluency       & 4.35  & 4.28  &  4.32 & \textbf{4.60} &  4.23  & 4.54  \\
& miss       & 22  & 16  &  15 & 11 &  14  & \textbf{9}  \\
& wrong       & 9  & 5  &  7 & 10 &  6  & \textbf{4}  \\
\end{tabular}
}
\caption{\label{tb:lmaug}\small Ablation studies for the impact of difficulty metrics on data, text or both (joint) with normalized scores including length (\textbf{L}), rarity (\textbf{R}), Damerau-Levenshtein Distance (\textbf{DLD}), plain edit distance (\textbf{PED}), and the proposed soft edit distance (\textbf{SED}). \textbf{Plain} means no curriculum learning techniques are added. All scores are computed based on the E2E corpus, consisting of both performance (BLEU) and the \textbf{average bin size}. Each \emph{bin} is defined by the number of training samples with the same difficulty scores.   
}
\end{table}


On Table~\ref{fig:step}, we observe that  \emph{soft edit distance} (SED) yields the best performance, outperforming a model that does not use curriculum learning by as much as $2.42$ BLEU.
It also outperforms all other metrics by roughly $1$ BLEU.
In general, we see that models perform better on \emph{joint} and \emph{text} than on \emph{data}.
This correlates to how a difficulty function is related to the \emph{average bin sizes} of scores it generates.
We see that for models that distinguish samples in a more defined manner, it will have a smaller average bin size where probability of having more difficult samples at every confidence threshold is lower.
From this, we see that \emph{length} and \emph{DLD} have larger average bin sizes across its difficulty scores and this makes samples less distinguishable from one another. 
Thus, they result in the smallest improvement over \emph{plain}. 
We show reordered samples in Table~\ref{tb:samples} for all difficulty metrics computed \emph{jointly} on data and text. We include length (\textbf{L}), rarity (\textbf{R}),  Damerau-Levenshtein Distance (\textbf{DLD}), and the proposed soft edit distance (\textbf{SED}).
 

On the other hand, we also justify the use of weighting for edit operation where \emph{PED}, which is the ``hard'' variant of \emph{SED} that does not weight edit operations like \emph{SED}, is shown to be far inferior to that of \emph{SED}.
The score margin comes up to $2.81$ BLEU.
Moreover, we further examine the difference in sample orders and observe that \emph{SED} yields more intuitive and better sample ordering as opposed to other metrics.

\paragraph{Human Evaluation.}
For \emph{human evaluation}, three annotators are instructed to evaluate 100 samples from the \emph{joint} variant to see (1) if the text is \emph{fluent} (score 0-5 with 5 being fully fluent), (2) if it \emph{misses} information contained in the source data and (3) if it includes \emph{wrong} information. 
These scores are averaged and presented in Table~\ref{tb:lmaug}.





\paragraph{On Training Speed.}

We define speed by the number of updates it takes to reach a performance plateau.
On Figure~\ref{fig:step}, the speedup is measured by the difference between the vertical bars. 
It can be observed that curriculum learning reduces the training steps to converge, where it consists of up to 38.7\% of the total updates for the same model without curriculum learning (on E2E).
Further, we see that the use of curriculum learning yields slightly worse performance in the initial training steps, but rise to a higher score and flattens as it converges.


\section{Conclusion}
\label{sec:conclusion}

To conclude, we show that the sample order does indeed matter when taking into account model competence during training.
Further, we demonstrate that the proposed metrics are effective in speeding up model convergence. Given that curriculum learning can be combined with pretty much any neural architecture, we recommend the use of curriculum learning for data-to-text generation.  
We believe this work offers insights into the annotation process of data with text labels where reduced number of labels are needed~\cite{hong2019improving,de2018generating,zhuang2017neobility,chang2020unsupervised,wiehr2020safe,shen2020neural,chang2020dart,su2020moviechats,chang2021neural,chang2021jointly}.

\section*{Acknowledgements}
This research was funded in part by the German Research Foundation (DFG) as part of SFB 248 ``Foundations of Perspicuous Software Systems''. We sincerely thank the anonymous reviewers for their insightful comments that helped us to improve this paper.

\bibliography{eacl2021,anthology}

\begin{thebibliography}{34}
\expandafter\ifx\csname natexlab\endcsname\relax\def\natexlab#1{#1}\fi

\bibitem[{Barzilay and Lapata(2005)}]{barzilay-lapata-2005-modeling}
Regina Barzilay and Mirella Lapata. 2005.
\newblock \href {https://doi.org/10.3115/1219840.1219858} {Modeling local
  coherence: An entity-based approach}.
\newblock In \emph{Proceedings of the 43rd Annual Meeting of the Association
  for Computational Linguistics ({ACL}{'}05)}, pages 141--148, Ann Arbor,
  Michigan. Association for Computational Linguistics.

\bibitem[{Bengio et~al.(2009)Bengio, Louradour, Collobert, and
  Weston}]{bengio2009curriculum}
Yoshua Bengio, J{\'e}r{\^o}me Louradour, Ronan Collobert, and Jason Weston.
  2009.
\newblock Curriculum learning.
\newblock In \emph{Proceedings of the 26th annual international conference on
  machine learning}, pages 41--48.

\bibitem[{Brill and Moore(2000{\natexlab{a}})}]{brill2000improved}
Eric Brill and Robert~C Moore. 2000{\natexlab{a}}.
\newblock An improved error model for noisy channel spelling correction.
\newblock In \emph{Proceedings of the 38th annual meeting of the association
  for computational linguistics}, pages 286--293.

\bibitem[{Brill and Moore(2000{\natexlab{b}})}]{brill-moore-2000-improved}
Eric Brill and Robert~C. Moore. 2000{\natexlab{b}}.
\newblock \href {https://doi.org/10.3115/1075218.1075255} {An improved error
  model for noisy channel spelling correction}.
\newblock In \emph{Proceedings of the 38th Annual Meeting of the Association
  for Computational Linguistics}, pages 286--293, Hong Kong. Association for
  Computational Linguistics.

\bibitem[{Chang et~al.(2020{\natexlab{a}})Chang, Adelani, Shen, and
  Demberg}]{chang2020unsupervised}
Ernie Chang, David Adelani, Xiaoyu Shen, and Vera Demberg. 2020{\natexlab{a}}.
\newblock Unsupervised pidgin text generation by pivoting english data and
  self-training.
\newblock In \emph{In Proceedings of Workshop at ICLR}.

\bibitem[{Chang et~al.(2020{\natexlab{b}})Chang, Caplinger, Marin, Shen, and
  Demberg}]{chang2020dart}
Ernie Chang, Jeriah Caplinger, Alex Marin, Xiaoyu Shen, and Vera Demberg.
  2020{\natexlab{b}}.
\newblock Dart: A lightweight quality-suggestive data-to-text annotation tool.
\newblock In \emph{COLING 2020}, pages 12--17.

\bibitem[{Chang et~al.(2021{\natexlab{a}})Chang, Demberg, and
  Marin}]{chang2021jointly}
Ernie Chang, Vera Demberg, and Alex Marin. 2021{\natexlab{a}}.
\newblock Jointly improving language understanding and generation with
  quality-weighted weak supervision of automatic labeling.
\newblock In \emph{EACL 2021}.

\bibitem[{Chang et~al.(2021{\natexlab{b}})Chang, Shen, Zhu, Demberg, and
  Su}]{chang2021neural}
Ernie Chang, Xiaoyu Shen, Dawei Zhu, Vera Demberg, and Hui. Su.
  2021{\natexlab{b}}.
\newblock Neural data-to-text generation with lm-based text augmentation.
\newblock \emph{EACL 2021}.

\bibitem[{Colin et~al.(2016)Colin, Gardent, M{'}rabet, Narayan, and
  Perez-Beltrachini}]{colin-etal-2016-webnlg}
Emilie Colin, Claire Gardent, Yassine M{'}rabet, Shashi Narayan, and Laura
  Perez-Beltrachini. 2016.
\newblock \href {https://doi.org/10.18653/v1/W16-6626} {The {W}eb{NLG}
  challenge: Generating text from {DBP}edia data}.
\newblock In \emph{Proceedings of the 9th International Natural Language
  Generation conference}, pages 163--167, Edinburgh, UK. Association for
  Computational Linguistics.

\bibitem[{Damerau(1964)}]{damerau1964technique}
Fred~J Damerau. 1964.
\newblock A technique for computer detection and correction of spelling errors.
\newblock \emph{Communications of the ACM}, 7(3):171--176.

\bibitem[{Elman(1993)}]{elman1993learning}
Jeffrey~L Elman. 1993.
\newblock Learning and development in neural networks: The importance of
  starting small.
\newblock \emph{Cognition}, 48(1):71--99.

\bibitem[{Gu et~al.(2019)Gu, Wang, and Zhao}]{gu2019levenshtein}
Jiatao Gu, Changhan Wang, and Junbo Zhao. 2019.
\newblock Levenshtein transformer.
\newblock In \emph{Advances in Neural Information Processing Systems}, pages
  11181--11191.

\bibitem[{Haffari(2009)}]{haffari2009machine}
Gholam~Reza Haffari. 2009.
\newblock \emph{Machine learning approaches for dealing with limited bilingual
  data in statistical machine translation}.
\newblock Ph.D. thesis, School of Computing Science-Simon Fraser University.

\bibitem[{Hong et~al.(2019)Hong, Chang, and Demberg}]{hong2019improving}
Xudong Hong, Ernie Chang, and Vera Demberg. 2019.
\newblock Improving language generation from feature-rich tree-structured data
  with relational graph convolutional encoders.
\newblock In \emph{Proceedings of the 2nd Workshop on Multilingual Surface
  Realisation (MSR 2019)}, pages 75--80.

\bibitem[{Kocmi and Bojar(2017)}]{kocmi2017curriculum}
Tom Kocmi and Ond{\v{r}}ej Bojar. 2017.
\newblock Curriculum learning and minibatch bucketing in neural machine
  translation.
\newblock In \emph{Proceedings of the International Conference Recent Advances
  in Natural Language Processing, RANLP 2017}, pages 379--386.

\bibitem[{Krueger and Dayan(2009)}]{krueger2009flexible}
Kai~A Krueger and Peter Dayan. 2009.
\newblock Flexible shaping: How learning in small steps helps.
\newblock \emph{Cognition}, 110(3):380--394.

\bibitem[{Lebret et~al.(2016)Lebret, Grangier, and Auli}]{lebret2016neural}
R{\'e}mi Lebret, David Grangier, and Michael Auli. 2016.
\newblock Neural text generation from structured data with application to the
  biography domain.
\newblock In \emph{Proceedings of the 2016 Conference on Empirical Methods in
  Natural Language Processing}, pages 1203--1213.

\bibitem[{Levenshtein(1966)}]{levenshtein1966binary}
Vladimir~I Levenshtein. 1966.
\newblock Binary codes capable of correcting deletions, insertions, and
  reversals.
\newblock In \emph{Soviet physics doklady}, volume~10, pages 707--710.

\bibitem[{Liu et~al.(2020)Liu, Lai, Wong, and Chao}]{liu2020norm}
Xuebo Liu, Houtim Lai, Derek~F Wong, and Lidia~S Chao. 2020.
\newblock Norm-based curriculum learning for neural machine translation.
\newblock \emph{arXiv preprint arXiv:2006.02014}.

\bibitem[{Novikova et~al.(2017)Novikova, Du{\v{s}}ek, and
  Rieser}]{novikova2017e2e}
Jekaterina Novikova, Ond{\v{r}}ej Du{\v{s}}ek, and Verena Rieser. 2017.
\newblock The e2e dataset: New challenges for end-to-end generation.
\newblock In \emph{Proceedings of the 18th Annual SIGdial Meeting on Discourse
  and Dialogue}, pages 201--206.

\bibitem[{Parikh et~al.(2020)Parikh, Wang, Gehrmann, Faruqui, Dhingra, Yang,
  and Das}]{parikh2020totto}
Ankur~P Parikh, Xuezhi Wang, Sebastian Gehrmann, Manaal Faruqui, Bhuwan
  Dhingra, Diyi Yang, and Dipanjan Das. 2020.
\newblock Totto: A controlled table-to-text generation dataset.
\newblock \emph{arXiv preprint arXiv:2004.14373}.

\bibitem[{Paszke et~al.(2019)Paszke, Gross, Massa, Lerer, Bradbury, Chanan,
  Killeen, Lin, Gimelshein, Antiga et~al.}]{pytorch}
Adam Paszke, Sam Gross, Francisco Massa, Adam Lerer, James Bradbury, Gregory
  Chanan, Trevor Killeen, Zeming Lin, Natalia Gimelshein, Luca Antiga, et~al.
  2019.
\newblock Pytorch: An imperative style, high-performance deep learning library.
\newblock In \emph{Advances in Neural Information Processing Systems}, pages
  8024--8035.

\bibitem[{Platanios et~al.(2019)Platanios, Stretcu, Neubig, Poczos, and
  Mitchell}]{Platanios2019}
Emmanouil~Antonios Platanios, Otilia Stretcu, Graham Neubig, Barnabas Poczos,
  and Tom~M. Mitchell. 2019.
\newblock \href {https://doi.org/10.18653/v1/n19-1119} {{Competence-based
  curriculum learning for neural machine translation}}.
\newblock \emph{NAACL HLT 2019 - 2019 Conference of the North American Chapter
  of the Association for Computational Linguistics: Human Language Technologies
  - Proceedings of the Conference}, 1:1162--1172.

\bibitem[{Plunkett and Marchman(1993)}]{plunkett1993rote}
Kim Plunkett and Virginia Marchman. 1993.
\newblock From rote learning to system building: Acquiring verb morphology in
  children and connectionist nets.
\newblock \emph{Cognition}, 48(1):21--69.

\bibitem[{Reiter and Dale(2000)}]{reiter2000building}
Ehud Reiter and Robert Dale. 2000.
\newblock \emph{Building natural language generation systems}.
\newblock Cambridge university press.

\bibitem[{Shen et~al.(2020)Shen, Chang, Su, Zhou, and Klakow}]{shen2020neural}
Xiaoyu Shen, Ernie Chang, Hui Su, Jie Zhou, and Dietrich Klakow. 2020.
\newblock Neural data-to-text generation via jointly learning the segmentation
  and correspondence.
\newblock In \emph{ACL 2020}.
  https://www.aclweb.org/anthology/2020.acl-main.641.pdf.

\bibitem[{de~Souza et~al.(2018)de~Souza, Kozielski, Mathur, Chang, Guerini,
  Negri, Turchi, and Matusov}]{de2018generating}
Jos{\'e}~GC de~Souza, Michael Kozielski, Prashant Mathur, Ernie Chang, Marco
  Guerini, Matteo Negri, Marco Turchi, and Evgeny Matusov. 2018.
\newblock Generating e-commerce product titles and predicting their quality.
\newblock In \emph{INLG}, pages 233--243.

\bibitem[{Su et~al.(2020)Su, Shen, Xiao, Zhang, Chang, Zhang, Niu, and
  Zhou}]{su2020moviechats}
Hui Su, Xiaoyu Shen, Zhou Xiao, Zheng Zhang, Ernie Chang, Cheng Zhang, Cheng
  Niu, and Jie Zhou. 2020.
\newblock Moviechats: Chat like humans in a closed domain.
\newblock In \emph{EMNLP 2020}, pages 6605--6619.

\bibitem[{Wiehr et~al.(2020)Wiehr, Hirsch, Daiber, Kruger, Kovtunova,
  Borgwardt, Chang, Demberg, Steinmetz, and Jorg}]{wiehr2020safe}
Frederik Wiehr, Anke Hirsch, Florian Daiber, Antonio Kruger, Alisa Kovtunova,
  Stefan Borgwardt, Ernie Chang, Vera Demberg, Marcel Steinmetz, and Hoffmann
  Jorg. 2020.
\newblock Safe handover in mixed-initiative control for cyber-physical systems.
\newblock In Proceedings of Workshop at CHI.

\bibitem[{Wiseman et~al.(2017)Wiseman, Shieber, and
  Rush}]{wiseman2017challenges}
Sam Wiseman, Stuart Shieber, and Alexander Rush. 2017.
\newblock Challenges in data-to-document generation.
\newblock In \emph{Proceedings of the 2017 Conference on Empirical Methods in
  Natural Language Processing}, pages 2253--2263.

\bibitem[{Zhang et~al.(2018)Zhang, Kumar, Khayrallah, Murray, Gwinnup,
  Martindale, McNamee, Duh, and Carpuat}]{zhang2018empirical}
Xuan Zhang, Gaurav Kumar, Huda Khayrallah, Kenton Murray, Jeremy Gwinnup,
  Marianna~J Martindale, Paul McNamee, Kevin Duh, and Marine Carpuat. 2018.
\newblock An empirical exploration of curriculum learning for neural machine
  translation.
\newblock \emph{arXiv preprint arXiv:1811.00739}.

\bibitem[{Zhang et~al.(2019)Zhang, Shapiro, Kumar, McNamee, Carpuat, and
  Duh}]{zhang2019curriculum}
Xuan Zhang, Pamela Shapiro, Gaurav Kumar, Paul McNamee, Marine Carpuat, and
  Kevin Duh. 2019.
\newblock Curriculum learning for domain adaptation in neural machine
  translation.
\newblock In \emph{Proceedings of the 2019 Conference of the North American
  Chapter of the Association for Computational Linguistics: Human Language
  Technologies, Volume 1 (Long and Short Papers)}, pages 1903--1915.

\bibitem[{Zhou et~al.(2020)Zhou, Yang, Wong, Wan, and
  Chao}]{zhou2020uncertainty}
Yikai Zhou, Baosong Yang, Derek~F Wong, Yu~Wan, and Lidia~S Chao. 2020.
\newblock Uncertainty-aware curriculum learning for neural machine translation.
\newblock In \emph{Proceedings of the 58th Annual Meeting of the Association
  for Computational Linguistics}, pages 6934--6944.

\bibitem[{Zhuang and Chang(2017)}]{zhuang2017neobility}
WenLi Zhuang and Ernie Chang. 2017.
\newblock Neobility at semeval-2017 task 1: An attention-based sentence
  similarity model.
\newblock In \emph{In Proceedings of SemEval-2017 at ACL 2017.}

\end{thebibliography}


\begin{thebibliography}{29}
\expandafter\ifx\csname natexlab\endcsname\relax\def\natexlab#1{#1}\fi

\bibitem[{Ambati(2012)}]{Ambati:2012}
Vamshi Ambati. 2012.
\newblock \emph{Active Learning and Crowdsourcing for Machine Translation in
  Low Resource Scenarios}.
\newblock Ph.D. thesis, Pittsburgh, PA, USA.
\newblock AAI3528171.

\bibitem[{Bahdanau et~al.(2015)Bahdanau, Cho, and Bengio}]{Bahdanau:2014}
Dzmitry Bahdanau, Kyunghyun Cho, and Yoshua Bengio. 2015.
\newblock \href {https://arxiv.org/pdf/1409.0473} {{Neural Machine Translation
  by Jointly Learning to Align and Translate}}.
\newblock In \emph{International Conference on Learning Representations}.

\bibitem[{Bawden et~al.(2018)Bawden, Sennrich, Birch, and
  Haddow}]{bawden2018discourse}
Rachel Bawden, Rico Sennrich, Alexandra Birch, and Barry Haddow. 2018.
\newblock \href {http://www.aclweb.org/anthology/N18-1118} {Evaluating
  discourse phenomena in neural machine translation}.
\newblock In \emph{Proceedings of the 2018 Conference of the North American
  Chapter of the Association for Computational Linguistics: Human Language
  Technologies, Volume 1 (Long Papers)}, pages 1304--1313, New Orleans,
  Louisiana. Association for Computational Linguistics.

\bibitem[{Bengio et~al.(2009)Bengio, Louradour, Collobert, and
  Weston}]{Bengio:2009}
Yoshua Bengio, J{\'e}r\^{o}me Louradour, Ronan Collobert, and Jason Weston.
  2009.
\newblock \href {https://doi.org/10.1145/1553374.1553380} {Curriculum
  learning}.
\newblock In \emph{Proceedings of the 26th Annual International Conference on
  Machine Learning}, ICML '09, pages 41--48, New York, NY, USA. ACM.

\bibitem[{Bloodgood and Callison-Burch(2010)}]{Bloodgood:2010}
Michael Bloodgood and Chris Callison-Burch. 2010.
\newblock Bucking the trend: Large-scale cost-focused active learning for
  statistical machine translation.
\newblock In \emph{ACL}.

\bibitem[{Bojar et~al.(2017{\natexlab{a}})Bojar, Chatterjee, Federmann, Graham,
  Haddow, Huang, Huck, Koehn, Liu, Logacheva et~al.}]{bojar2017findings}
Ond{\v{r}}ej Bojar, Rajen Chatterjee, Christian Federmann, Yvette Graham, Barry
  Haddow, Shujian Huang, Matthias Huck, Philipp Koehn, Qun Liu, Varvara
  Logacheva, et~al. 2017{\natexlab{a}}.
\newblock Findings of the 2017 conference on machine translation (wmt17).
\newblock In \emph{Proceedings of the Second Conference on Machine
  Translation}, pages 169--214.

\bibitem[{Bojar et~al.(2017{\natexlab{b}})Bojar, Helcl, Kocmi, Libovick\'{y},
  and Musil}]{Bojar:2017}
Ond\v{r}ej Bojar, Jind\v{r}ich Helcl, Tom Kocmi, Jind\v{r}ich Libovick\'{y},
  and Tom\'{a}\v{s} Musil. 2017{\natexlab{b}}.
\newblock \href {http://www.aclweb.org/anthology/W17-4757} {{Results of the
  WMT17 Neural MT Training Task}}.
\newblock In \emph{Proceedings of the Second Conference on Machine Translation,
  Volume 2: Shared Task Papers}, pages 525--533. Association for Computational
  Linguistics.

\bibitem[{Chen et~al.(2018)Chen, Firat, Bapna, Johnson, Macherey, Foster,
  Jones, Schuster, Shazeer, Parmar, Vaswani, Uszkoreit, Kaiser, Chen, Wu, and
  Hughes}]{Chen:2018}
Mia~Xu Chen, Orhan Firat, Ankur Bapna, Melvin Johnson, Wolfgang Macherey,
  George Foster, Llion Jones, Mike Schuster, Noam Shazeer, Niki Parmar, Ashish
  Vaswani, Jakob Uszkoreit, Lukasz Kaiser, Zhifeng Chen, Yonghui Wu, and
  Macduff Hughes. 2018.
\newblock \href {http://aclweb.org/anthology/P18-1008} {{The Best of Both
  Worlds: Combining Recent Advances in Neural Machine Translation}}.
\newblock In \emph{Proceedings of the 56th Annual Meeting of the Association
  for Computational Linguistics (Long Papers)}, pages 76--86. Association for
  Computational Linguistics.

\bibitem[{Crego et~al.(2016)Crego, Kim, Klein, Rebollo, Yang, Senellart,
  Akhanov, Brunelle, Coquard, Deng, Enoue, Geiss, Johanson, Khalsa, Khiari, Ko,
  Kobus, Lorieux, Martins, Nguyen, Priori, Riccardi, Segal, Servan, Tiquet,
  Wang, Yang, Zhang, Zhou, and Zoldan}]{Crego:2016}
Josep~Maria Crego, Jungi Kim, Guillaume Klein, Anabel Rebollo, Kathy Yang, Jean
  Senellart, Egor Akhanov, Patrice Brunelle, Aurelien Coquard, Yongchao Deng,
  Satoshi Enoue, Chiyo Geiss, Joshua Johanson, Ardas Khalsa, Raoum Khiari,
  Byeongil Ko, Catherine Kobus, Jean Lorieux, Leidiana Martins, Dang-Chuan
  Nguyen, Alexandra Priori, Thomas Riccardi, Natalia Segal, Christophe Servan,
  Cyril Tiquet, Bo~Wang, Jin Yang, Dakun Zhang, Jing Zhou, and Peter Zoldan.
  2016.
\newblock \href {https://arxiv.org/abs/1610.05540} {{SYSTRAN's Pure Neural
  Machine Translation Systems}}.
\newblock \emph{CoRR}, abs/1610.05540.

\bibitem[{Elman(1993)}]{elman1993learning}
Jeffrey~L Elman. 1993.
\newblock Learning and development in neural networks: The importance of
  starting small.
\newblock \emph{Cognition}, 48(1):71--99.

\bibitem[{Haffari et~al.(2009)Haffari, Roy, and Sarkar}]{Haffari:2009}
Gholamreza Haffari, Maxim Roy, and Anoop Sarkar. 2009.
\newblock \href {http://dl.acm.org/citation.cfm?id=1620754.1620815} {Active
  learning for statistical phrase-based machine translation}.
\newblock In \emph{Proceedings of Human Language Technologies: The 2009 Annual
  Conference of the North American Chapter of the Association for Computational
  Linguistics}, NAACL '09, pages 415--423, Stroudsburg, PA, USA. Association
  for Computational Linguistics.

\bibitem[{Kalchbrenner and Blunsom(2013)}]{kalchbrenner2013recurrent}
Nal Kalchbrenner and Phil Blunsom. 2013.
\newblock Recurrent continuous translation models.
\newblock In \emph{Proceedings of the 2013 Conference on Empirical Methods in
  Natural Language Processing}, pages 1700--1709.

\bibitem[{Kocmi and Bojar(2017)}]{Kocmi:2017}
Tom Kocmi and Ond{\v{r}}ej Bojar. 2017.
\newblock \href {https://doi.org/10.26615/978-954-452-049-6_050} {{Curriculum
  Learning and Minibatch Bucketing in Neural Machine Translation}}.
\newblock In \emph{Proceedings of the International Conference Recent Advances
  in Natural Language Processing}, pages 379--386.

\bibitem[{Krueger and Dayan(2009)}]{krueger2009flexible}
Kai~A Krueger and Peter Dayan. 2009.
\newblock Flexible shaping: How learning in small steps helps.
\newblock \emph{Cognition}, 110(3):380--394.

\bibitem[{Liu et~al.(2018)Liu, Lu, and Neubig}]{liu2018homographs}
Frederick Liu, Han Lu, and Graham Neubig. 2018.
\newblock \href {http://www.aclweb.org/anthology/N18-1121} {Handling homographs
  in neural machine translation}.
\newblock In \emph{Proceedings of the 2018 Conference of the North American
  Chapter of the Association for Computational Linguistics: Human Language
  Technologies, Volume 1 (Long Papers)}, pages 1336--1345, New Orleans,
  Louisiana. Association for Computational Linguistics.

\bibitem[{Platanios(2018)}]{Platanios:2018b}
Emmanouil~A. Platanios. 2018.
\newblock {TensorFlow Scala}.
\newblock \url{https://github.com/eaplatanios/tensorflow_scala}.

\bibitem[{Platanios et~al.(2018)Platanios, Sachan, Neubig, and
  Mitchell}]{Platanios:2018}
Emmanouil~Antonios Platanios, Mrinmaya Sachan, Graham Neubig, and Tom Mitchell.
  2018.
\newblock \href {https://arxiv.org/abs/1808.08493} {{Contextual Parameter
  Generation for Universal Neural Machine Translation}}.
\newblock In \emph{Conference on Empirical Methods in Natural Language
  Processing (EMNLP)}, Brussels, Belgium.

\bibitem[{Popel and Bojar(2018)}]{Popel:2018}
Martin Popel and Ond{\v{r}}ej Bojar. 2018.
\newblock \href
  {https://content.sciendo.com/view/journals/pralin/110/1/article-p43.xml}
  {{Training Tips for the Transformer Model}}.
\newblock \emph{The Prague Bulletin of Mathematical Linguistics},
  110(1):43--70.

\bibitem[{Reddi et~al.(2018)Reddi, Kale, and Kumar}]{Reddi:2018}
Sashank~J. Reddi, Satyen Kale, and Sanjiv Kumar. 2018.
\newblock \href {https://openreview.net/forum?id=ryQu7f-RZ} {{On the
  Convergence of Adam and Beyond}}.
\newblock In \emph{International Conference on Learning Representations}.

\bibitem[{Schapire(1999)}]{Schapire:1999}
Robert~E. Schapire. 1999.
\newblock \href {http://dl.acm.org/citation.cfm?id=1624312.1624417} {A brief
  introduction to boosting}.
\newblock In \emph{Proceedings of the 16th International Joint Conference on
  Artificial Intelligence - Volume 2}, IJCAI'99, pages 1401--1406, San
  Francisco, CA, USA. Morgan Kaufmann Publishers Inc.

\bibitem[{Sennrich et~al.(2016)Sennrich, Haddow, and Birch}]{Sennrich:2016a}
Rico Sennrich, Barry Haddow, and Alexandra Birch. 2016.
\newblock \href {http://www.aclweb.org/anthology/P16-1162} {{Neural Machine
  Translation of Rare Words with Subword Units}}.
\newblock In \emph{Proceedings of the 54th Annual Meeting of the Association
  for Computational Linguistics}, pages 1715--1725.

\bibitem[{Settles and Craven(2008)}]{settles2008activelearning}
Burr Settles and Mark Craven. 2008.
\newblock \href {http://www.aclweb.org/anthology/D08-1112} {An analysis of
  active learning strategies for sequence labeling tasks}.
\newblock In \emph{Proceedings of the 2008 Conference on Empirical Methods in
  Natural Language Processing}, pages 1070--1079, Honolulu, Hawaii. Association
  for Computational Linguistics.

\bibitem[{Shazeer and Stern(2018)}]{Shazeer:2018}
Noam Shazeer and Mitchell Stern. 2018.
\newblock \href {http://proceedings.mlr.press/v80/shazeer18a.html} {{Adafactor:
  Adaptive Learning Rates with Sublinear Memory Cost}}.
\newblock In \emph{Proceedings of the 35th International Conference on Machine
  Learning}, volume~80 of \emph{Proceedings of Machine Learning Research},
  pages 4596--4604, Stockholmsmässan, Stockholm Sweden. PMLR.

\bibitem[{Sutskever et~al.(2014)Sutskever, Vinyals, and
  Le}]{sutskever2014sequence}
Ilya Sutskever, Oriol Vinyals, and Quoc~V Le. 2014.
\newblock Sequence to sequence learning with neural networks.
\newblock In \emph{Advances in neural information processing systems}, pages
  3104--3112.

\bibitem[{Vaswani et~al.(2017)Vaswani, Shazeer, Parmar, Uszkoreit, Jones,
  Gomez, Kaiser, and Polosukhin}]{Vaswani:2017}
Ashish Vaswani, Noam Shazeer, Niki Parmar, Jakob Uszkoreit, Llion Jones,
  Aidan~N Gomez, Lukasz Kaiser, and Illia Polosukhin. 2017.
\newblock \href
  {http://papers.nips.cc/paper/7181-attention-is-all-you-need.pdf} {{Attention
  is All you Need}}.
\newblock In \emph{Advances in Neural Information Processing Systems}, pages
  5998--6008.

\bibitem[{Wu et~al.(2016)Wu, Schuster, Chen, Le, Norouzi, Macherey, Krikun,
  Cao, Gao, Macherey, Klingner, Shah, Johnson, Liu, Łukasz Kaiser, Gouws,
  Kato, Kudo, Kazawa, Stevens, Kurian, Patil, Wang, Young, Smith, Riesa,
  Rudnick, Vinyals, Corrado, Hughes, and Dean}]{Wu:2016}
Yonghui Wu, Mike Schuster, Zhifeng Chen, Quoc~V. Le, Mohammad Norouzi, Wolfgang
  Macherey, Maxim Krikun, Yuan Cao, Qin Gao, Klaus Macherey, Jeff Klingner,
  Apurva Shah, Melvin Johnson, Xiaobing Liu, Łukasz Kaiser, Stephan Gouws,
  Yoshikiyo Kato, Taku Kudo, Hideto Kazawa, Keith Stevens, George Kurian,
  Nishant Patil, Wei Wang, Cliff Young, Jason Smith, Jason Riesa, Alex Rudnick,
  Oriol Vinyals, Greg Corrado, Macduff Hughes, and Jeffrey Dean. 2016.
\newblock \href {https://arxiv.org/abs/1609.08144} {{Google's Neural Machine
  Translation System: Bridging the Gap between Human and Machine Translation}}.
\newblock \emph{CoRR}, abs/1609.08144.

\bibitem[{Zhang et~al.(2016)Zhang, Kim, Crego, and
  Senellart}]{zhang2016boosting}
Dakun Zhang, Jungi Kim, Josep Crego, and Jean Senellart. 2016.
\newblock Boosting neural machine translation.
\newblock \emph{arXiv preprint arXiv:1612.06138}.

\bibitem[{Zhang et~al.(2018)Zhang, Kumar, Khayrallah, Murray, Gwinnup,
  Martindale, McNamee, Duh, and Carpuat}]{Zhang:2018}
Xuan Zhang, Gaurav Kumar, Huda Khayrallah, Kenton Murray, Jeremy Gwinnup,
  Marianna~J Martindale, Paul McNamee, Kevin Duh, and Marine Carpuat. 2018.
\newblock \href {https://arxiv.org/abs/1811.00739} {{An Empirical Exploration
  of Curriculum Learning for Neural Machine Translation}}.
\newblock \emph{CoRR}, abs/1811.00739.

\bibitem[{Zou et~al.(2013)Zou, Socher, Cer, and Manning}]{Zou:2013}
Will~Y. Zou, Richard Socher, Daniel Cer, and Christopher~D. Manning. 2013.
\newblock \href {http://aclweb.org/anthology/D13-1141} {{Bilingual Word
  Embeddings for Phrase-Based Machine Translation}}.
\newblock In \emph{Proceedings of the 2013 Conference on Empirical Methods in
  Natural Language Processing}, pages 1393--1398. Association for Computational
  Linguistics.

\end{thebibliography}
\bibliographystyle{acl_natbib}

\end{document}